\documentclass[
]{ceurart}

\usepackage{listings}
\usepackage{todonotes}
\usepackage[utf8]{inputenc}
\usepackage[T1]{fontenc}
\usepackage{placeins}
\begin{document}

\copyrightyear{2024}
\copyrightclause{Copyright for this paper by its authors.
  Use permitted under Creative Commons License Attribution 4.0
  International (CC BY 4.0).}

\conference{CLEF 2024: Conference and Labs of the Evaluation Forum, September 09–12, 2024, Grenoble, France}

\title{OpenFact at CheckThat! 2024: Combining Multiple Attack Methods for Effective Adversarial Text Generation}

\title[mode=sub]{Notebook for the CheckThat! Lab at CLEF 2024}

\author[1]{Włodzimierz Lewoniewski}[%
orcid=0000-0002-0163-5492,
email=wlodzimierz.lewoniewski@ue.poznan.pl,
]
\address[1]{Department of Information Systems, Poznań University of Economics and Business,
	Al. Niepodległości 10, 61-875 Poznań, Poland}

\author[1]{Piotr Stolarski}[%
orcid=0000-0001-7076-2316,
email=piotr.stolarski@ue.poznan.pl,
]

\author[1]{Milena Stróżyna}[%
orcid=0000-0001-7603-7369,
email=milena.strozyna@ue.poznan.pl,
]
\cormark[1]

\author[1]{Elzbieta Lewańska}[%
orcid=0000-0002-0865-803X,
email=elzbieta.lewanska@ue.poznan.pl,
]

\author[1]{Aleksandra Wojewoda}[%
orcid=0000-0001-6160-1617,
email=aleksandra.wojewoda@ue.poznan.pl,
]

\author[1]{Ewelina Księżniak}[%
orcid=0000-0003-1953-8014,
email=ewelina.ksiezniak@ue.poznan.pl,
]

\author[1]{Marcin Sawiński}[%
orcid=0000-0002-1226-4850,
email=marcin.sawinski@ue.poznan.pl,
]

\cortext[1]{Corresponding author.}
\fntext[1]{These authors contributed equally.}

\begin{abstract}
  This paper presents the experiments and results for the CheckThat! Lab at CLEF 2024 Task 6: Robustness of Credibility Assessment with Adversarial Examples (InCrediblAE). The primary objective of this task was to generate adversarial examples in five problem domains in order to evaluate the robustness of widely used text classification methods (fine-tuned BERT, BiLSTM, and RoBERTa) when applied to credibility assessment issues.
  This study explores the application of ensemble learning to enhance adversarial attacks on natural language processing (NLP) models. We systematically tested and refined several adversarial attack methods, including BERT-Attack, Genetic algorithms, TextFooler, and CLARE, on five datasets across various misinformation tasks. By developing modified versions of BERT-Attack and hybrid methods, we achieved significant improvements in attack effectiveness. Our results demonstrate the potential of modification and combining multiple methods to create more sophisticated and effective adversarial attack strategies, contributing to the development of more robust and secure systems.
\end{abstract}


\maketitle

\section{Introduction}
\label{intro}
Robustness of credibility assessment is an interesting area of research that helps to understand the limitations of automatic classification methods used in variety of applications.
Therefore we began our study with a comprehensive literature review to understand the current state of adversarial attack methods. We systematically tested various adversarial attack techniques on BERT and BiLSTM classifiers \cite{przybyła2023verifying}, focusing on four misinformation tasks: Fake News Classification (FC), Hate Speech Detection (HN), Propaganda Recognition (PR), and Rumor Detection (RD). Previous work in this area demonstrated that BERT-Attack  \cite{li2020bertattack} and Genetic algorithms \cite{alzantot2018generating} often yield the best results depending on the task and victim model \cite{przybyła2023verifying}. Other methods, including DeepWordBug \cite{gao2018black} and SCPN \cite{iyyer2018adversarial}, also showed high effectiveness in specific scenarios.

Inspired by findings from previous works and the variety of plausible techniques, we decided to test and modify them in order to combine the most effective. We utilized our experiments on five datasets: C19, FC, HN, PR2, and RD. Utilizing the OpenAttack Python library \cite{zeng2020openattack}, we implemented and tested various adversarial attack methods. Our goal was to refine these methods and develop an optimized ensemble approach to adversarial attacks.

In our subsequent experiments, we modified the BERT-Attack  method to balance semantic preservation with attack success rate. By adjusting parameters and exploring the impact of increased substitute numbers, we developed a modified version of BERT-Attack, termed BAm. Additionally, we introduced a new method for selecting important words for replacement, leading to further enhancements in attack effectiveness.

To improve results, we integrated the Genetic algorithm for cases where initial modifications failed, creating a hybrid approach. We also experimented with a synonym replacement method based on word embeddings. Finally, we incorporated the CLARE method \cite{li2021contextualized}, which employs a unique mask-then-infill procedure using a pre-trained masked language model, to achieve superior attack success rates while preserving textual similarity and fluency.

Through this study, we aim to demonstrate the potential of ensemble learning in enhancing adversarial attacks on various models. Our findings can provide valuable insights into the development of more sophisticated and effective adversarial attack strategies, ultimately contributing to the resilience and security of various systems.

\section{Background}
\label{methodology}
The paper presents the experiments and results for the CheckThat! Lab at CLEF 2024 Task 6: Robustness of Credibility Assessment with Adversarial Examples (InCrediblAE) \cite{przybyla2024overview,barron2024overview}.  
The objective of this task was to evaluate the robustness of widely used text classification methods when applied to credibility assessment issues. 
There were three trained victim classifiers provided by task’s organizers: fine-tuned BERT, BiLSTM, and RoBERTa. 
The first two classifiers were made available as soon as the task was announced, so they were the foundation for our approaches and conducted experiments. 
The latter was made available only in the test phase and used to evaluate our methods.  
Moreover, five problem domains were defined, each being a binary classification task, and for each problem, a separate dataset was provided: 
\begin{itemize}
	\item Style-based news bias assessment (HN dataset),
	\item Propaganda detection (PR2 dataset),
 \item Fact checking (FC dataset),
	\item Rumour detection (RD dataset),
	\item COVID-19 misinformation detection (C19 dataset).
\end{itemize}

Our task was to create adversarial examples for each dataset. 
The adversarial examples should include small modifications in each dataset’s record that, on the one hand, would preserve the meaning (semantics) of the original text and, on the other hand, would change the victim classifier’s decision. 

To complete the task, we started with an in-depth analysis of the relevant literature on adversarial attacks on NLP classification and the BODEGA framework, as described in section \ref{sota}. 
In parallel, we analyzed the datasets provided by the task’s organizers (see below). 
In the next step, we applied the existing attack methods to the provided datasets and treated the received results as a baseline for our approach. 
The baseline attack methods encompassed those used by \cite{przybyła2023verifying}, namely BAE, BERT-ATTACK, DeepWordBug, Genetic, SemesePSO, PWWS, SCPN, TextFooler, implemented in the OpenAttack library (see section \ref{sota}).
Finally, we developed and tested own approaches to create adversarial examples for each dataset, which are described in detail in the following section. 
The received adversarial examples for each dataset were evaluated using the BODEGA score, consisting of three components: confusion score, semantic score and character score. 
Our aim was to maximize BODEGA.
In our experiments, we worked solely on the attack datasets that were provided for the task (the train and dev datasets were not used). 
Table \ref{tab:datasets} presents the basic statistics on the datasets used.

\begin{table*}
	\caption{Datasets statistics}
	\label{tab:datasets}
	\begin{tabular}{cccc}
		\toprule
		\multirow{2}{*}{Dataset} & \multicolumn{2}{c}{Label} & \multirow{2}{*}{Total} \\
		& 1 (non-credible) & 0 (credible) &  \\
		\midrule
			C19 & 280 & 315 & 595 \\
			FC & 205 & 200 & 405 \\
			HN & 202 & 198 & 400 \\
			PR2 & 123 & 293 & 416 \\
			RD & 136 & 279 & 415\\
		\bottomrule
	\end{tabular}
\end{table*}

The detailed description of datasets (apart from C19) is provided in \cite{przybyła2023verifying}. For each datasets, we modified the following elements to get the adversarial examples:
\begin{itemize}
	\item C19 dataset – a pair of texts – target claim and/or relevant evidence. The output label indicates whether the evidence supports the claim or refutes it.
	\item FC dataset - a pair of texts – target claim and/or relevant evidence. The output label indicates whether the evidence supports the claim or refutes it. The source of each evidence (Wikipedia post) was also provided, but this information was not used.
	\item 	HN dataset – news article. The label is assigned based on the overall bias of the source, assessed by independent journalists. A link to the article was also provided, but we did not use it. 
	\item PR2 dataset – sentences. The label indicates if sentences contain any propaganda instances or not. Information about the origin of the dataset (SemEval 2020 Task corpus) was also provided, but we did not use this information.
	\item 	RD dataset – a conversation between Twitter users, i.e., the original post and follow-ups from other Twitter users (Twitter threads converted to a flat feed of concatenated texts, containing the initial post and subsequent responses). The label indicates if the thread is a rumor or not.  A link to the Twitter thread was provided,  but it was not used. 
\end{itemize}	

The experiments were conducted on our own infrastructure, consisting of four NVIDIA GeForce RTX 2080 Ti GPU cards, each with 11 GB of VRAM memory and one NVIDIA GeForce RTX 4070 GPU card with 12 GB of VRAM memory.

\section{Related Work}
\label{sota}

Adversarial attacks are manipulations of input text designed to deceive machine learning models and cause them to make errors. Adversarial attacks often involve making subtle, human-imperceptible changes to the input, which can lead the model to produce incorrect or unexpected outputs. Those changes might include one or many of the following, exemplary scenarios: character-level (adding typos in the text, adding extra characters at the beginning/end of the sentence), word-level (replacing word with its synonym, adding semantically neutral words), sentence-level (paraprasing) \cite{xu_llm_2023, yang_fast_2024}. Adversarial attacks are used for ML-models tuning in order to increase their robustness \cite{dang_curious_2024, yang_fast_2024, xu_llm_2023, zhang_adversarial_2020}

The goal of the presented research was to create adversarial examples targeted at victim models that classify text instances in five different domains (style-based news bias assessment, propaganda detection, fact checking, rumour detection, and COVID-19 misinformation detection) described further in section \ref{methodology}. 

The methods presented in this paper have been evaluated within a framework provided in InCrediblAE, called BODEGA (Benchmark fOr aDversarial Example Generation in credibility Assessment) \cite{przybyła2023verifying} -- a framework for testing adversarial examples generation solutions. In general, there are two goal functions in the adversarial attack: (1) to maximise difference between the classes predicted by the classifier for the original and alternated instances; and (2) to maximise similarity between the original and alternated instances. The BODEGA measure ranges from 0 to 1, where 0 means that the original text has been completely altered, including a change in its meaning, and 1 means that the altered text has retained the original semantics, while having only a minimal edit distance from the original text.

The BODEGA is defined as follows:

\begin{equation}
\label{eq:bodega}
\text{BODEGA\_score}(x_i, x_i^*) = \text{Con\_score}(x_i, x_i^*) \times \text{Sem\_score}(x_i, x_i^*) \times \text{Char\_score}(x_i, x_i^*),
\end{equation}

where $x_i$ is original text instance and $x_i^*$ is modified text instance. There are three components of the BODEGA score. $Con\_score$ has 0 or 1 value (1 if the text instance alternation successfully caused classifier to change its decision, and 0 otherwise). $Sem\_score$ uses BLEURT \cite{sellam_bleurt_2020} -- a measure of semantic similarity between texts (ranges between 0 and 1, where 1 means that two text has the same semantic meaning). $Char\_score$ is based on Levenshtein distance and define as follows:

\begin{equation}
\text{Char\_score}(a, b) = 1 - \frac{\text{lev\_dist}(a, b)}{\max(|a|, |b|)}
\end{equation}

Final BODEGA score is calculated an average over BODEGA scores for all instances in the attack set (i.e. all adversarial attack instances).

To generate adversarial attack examples, two notable Python libraries are TextAttack \cite{morris2020textattack} and OpenAttack \cite{zeng2020openattack}.
TextAttack is a Python library designed to execute adversarial attacks in NLP by breaking them down into four components: goal functions, constraints, transformations, and search methods. Goal functions evaluate the success of an attack based on model outputs, while constraints ensure that perturbations meet specific requirements, such as minimum sentence encoding cosine similarity or maximum word embedding distance. Transformations generate perturbations through selected methods (for example word swaps or word embedding swaps), and search methods iteratively select the most promising perturbations. The library includes implementations of 16 adversarial attacks and provides access to 82 pre-trained models \cite{morris2020textattack}.
Similar to TextAttack, OpenAttack features a modular design for the swift implementation of various attack models. However, OpenAttack distinguishes itself by supporting more types of attacks, including sentence-level attacks, and by offering multilingual support for both English and Chinese, with the potential for additional languages. Additionally, OpenAttack enhances efficiency through parallel process execution, addressing the time-consuming nature of some attack models \cite{zeng2020openattack}.

Based on the literature and our own preliminary test, we identified methods available in TexAttack and/or OpenAttack that were selected to include in further experiments. Those were the following methods:
\begin{itemize}
	\item BAE  \cite{garg_bae_2020} -- BERT-based, masked model that uses word perturbation; BAE exchanges tokens with substitutes or adds new ones that are suitable in a given context;
	\item BERT-ATTACK \cite{li2020bertattack} -- BERT-based, masked model that uses word perturbation; BERT-ATTACK identifies tokens that most likely influence the victim's decision and exchanges them with substitutes;
	\item PWWS \cite{ren_generating_2019} --  uses word perturbation, uses WordNet in order to identify synonyms;
	\item Genetic \cite{alzantot2018generating} --  uses word perturbation; uses a genetic algorithm, where in each population, a number of different texts are generated, and those with the highest scores are selected for further replication;
	\item SememePSO \cite{yuan_zang_word-level_2020} --  uses word perturbation; uses Particle Swarm Optimisation (PSO), where the goal is to find the optimal posion of a set of different, modified texts (represented by particles) in the feature space; 
	\item TextFooler \cite{jin2020bert} --  uses word perturbation,It replaces tokens with their substitutes but considers the part of speech of the replaced token, thus preserving the original meaning of the sentence;
	\item CLARE \cite{li2021contextualized} --  masked model that uses word perturbation; CLARE also identifies tokens that most likely influence the victim's decision and exchanges them with its substitutes, but it allows the input of a new token at any position, not only at the exact same position as the replaced token;
	\item DeepWordBug \cite{gao2018black} -- uses char perturbation; DeepWordBug replaces characters so that the word is not recognizable by the victim classifier; the changes are usually not noticeable to humans;
	\item SCPN \cite{iyyer2018adversarial} -- uses sentence perturbation; uses paraphrasing of the whole text.
\end{itemize}

\section{Methodology}
\label{experiments}

In general our approach can be characterized as a special case of ensemble learning. We took the inspiration from this machine learning group of techniques. In case of ensemble learning it is assumed that employment of multiple learning algorithms will outperform any single algorithm used alone.

Although the idea of an ensemble is relatively simple it is more problematic on an implementational level. Especially it is hard to be deployed with respect to new application. In our case this new application is the adversarial attack domain. There are a number of difficulties to solve:

\begin{itemize}
\item Firstly, there is a variety of approaches that can be taken when it comes to the realization of the ensemble learning. It has to be decided which specific approach will fit best to the projected task.
\item Secondly, there is an abundant set of methods already used for the execution for the adversarial attack task. The decision has to be made how to create a subset for methods that should be used as one of the ensemble algorithms.
\end{itemize}

In our approach we have started from a literature review and then systematically tested some of the adversarial attack methods. One of the important  works in this area \cite{przybyła2023verifying} tested different adversarial attacks on the BERT and BiLSTM classifiers (victims) in four misinformation tasks: FC, HN, PR, RD. The result showed, that generally depending on the task and victim model the best results can be obtained using BERT-ATTACK \cite{li2020bertattack} and Genetic algorithm \cite{alzantot2018generating}. In some specific cases and measures the highest scores can be obtained using such approaches as TextFooler \cite{jin2020bert}, DeepWordBug \cite{gao2018black} and SCPN \cite{iyyer2018adversarial}. Therefore we decided to test proposed adversarial attacks on our five attack datasets: C19, FC, HN, PR, RD. Similarly to the mentioned work, we also used implementations of selected adversarial attack methods in the OpenAttack Python library \cite{openattack} in our first stages of experiments. Results are presented in the table \ref{tab:baseline}.

Next we decided to experiment with different settings and provided own modifications to the BERT-attacker implemented in OpenAttack library for better results. Due to the fact, that balance between semantic preservation and attack success rate can be regulated throughout a threshold of semantic similarity score of substituted word, we modified ''threshold\_pred\_score'' parameter from 0.3 to 0.2. Also we decided to check how twice-increased number of substitutes (from 36 to 72) will affect results. Those modificated version of the BERT-Attack we called \textbf{BAm}. The table \ref{tab:bam} presents a comparison of BERT-Attack with default parameters in OpenAttack library ('BA') and modified version ('BAm') on five datasets and three victims in BODEGA, success, semantic, character scores and queries number per example.

\begin{table}[]
	\caption{Comparison of BERT-Attack with default parameters in OpenAttack library ('BA') and modified version ('BAm') on five datasets and three victims in BODEGA, success, semantic, character scores and queries number per example.}
	\label{tab:bam}
	\begin{tabular}{ll|rr|rr|rr|rr|rr}
		\multicolumn{1}{c}{\multirow{2}{*}{\textbf{Task}}} & \multicolumn{1}{c}{\multirow{2}{*}{\textbf{Victim}}} & \multicolumn{2}{c}{\textbf{BODEGA}}                                & \multicolumn{2}{c}{\textbf{Success}}                               & \multicolumn{2}{c}{\textbf{Semantic}}                              & \multicolumn{2}{c}{\textbf{Character}}                             & \multicolumn{2}{c}{\textbf{Queries}}                               \\
		\multicolumn{1}{c}{}                               & \multicolumn{1}{c}{}                                 & \multicolumn{1}{c}{\textbf{BA}} & \multicolumn{1}{c}{\textbf{BAm}} & \multicolumn{1}{c}{\textbf{BA}} & \multicolumn{1}{c}{\textbf{BAm}} & \multicolumn{1}{c}{\textbf{BA}} & \multicolumn{1}{c}{\textbf{BAm}} & \multicolumn{1}{c}{\textbf{BA}} & \multicolumn{1}{c}{\textbf{BAm}} & \multicolumn{1}{c}{\textbf{BA}} & \multicolumn{1}{c}{\textbf{BAm}} \\ \hline
		C19                                                & BERT                                                 & 0.43                            & 0.45                             & 0.75                            & 0.78                             & 0.60                            & 0.60                             & 0.95                            & 0.95                             & 158.05                          & 243.60                           \\
		C19                                                & BiLSTM                                               & 0.49                            & 0.51                             & 0.85                            & 0.88                             & 0.61                            & 0.61                             & 0.95                            & 0.95                             & 127.06                          & 184.54                           \\
		C19                                                & RoBERTa                                             & 0.37                            & 0.41                             & 0.70                            & 0.78                             & 0.57                            & 0.57                             & 0.92                            & 0.93                             & 193.92                          & 294.01                           \\ \hline
		FC                                                 & BERT                                                 & 0.54                            & 0.54                             & 0.78                            & 0.78                             & 0.72                            & 0.72                             & 0.95                            & 0.95                             & 145.93                          & 145.93                           \\
		FC                                                 & BiLSTM                                               & 0.60                            & 0.62                             & 0.86                            & 0.89                             & 0.73                            & 0.72                             & 0.95                            & 0.95                             & 133.14                          & 182.64                           \\
		FC                                                 & RoBERTa                                             & 0.55                            & 0.59                             & 0.79                            & 0.84                             & 0.73                            & 0.73                             & 0.95                            & 0.95                             & 161.67                          & 213.51                           \\ \hline
		HN                                                 & BERT                                                 & 0.59                            & 0.61                             & 0.96                            & 0.99                             & 0.63                            & 0.63                             & 0.97                            & 0.98                             & 615.09                          & 723.45                           \\
		HN                                                 & BiLSTM                                               & 0.62                            & 0.63                             & 0.97                            & 0.99                             & 0.64                            & 0.64                             & 0.99                            & 0.99                             & 529.17                          & 542.83                           \\
		HN                                                 & RoBERTa                                             & 0.38                            & 0.42                            & 0.67                            & 0.74                             & 0.60                            & 0.60                             & 0.95                            & 0.95                             & 1744.08                         & 2689.37                          \\ \hline
		PR2                                                & BERT                                                 & 0.43                            & 0.49                             & 0.69                            & 0.80                             & 0.68                            & 0.67                             & 0.90                            & 0.90                             & 80.20                           & 113.68                           \\
		PR2                                                & BiLSTM                                               & 0.53                            & 0.55                             & 0.80                            & 0.85                             & 0.72                            & 0.71                             & 0.91                            & 0.91                             & 61.60                           & 87.81                            \\
		PR2                                                & RoBERTa                                             & 0.24                            & 0.27                             & 0.39                            & 0.46                             & 0.67                            & 0.66                             & 0.89                            & 0.89                             & 119.89                          & 200.59                           \\ \hline
		RD                                                 & BERT                                                 & 0.19                            & 0.22                             & 0.43                            & 0.50                             & 0.45                            & 0.45                             & 0.96                            & 0.96                             & 763.34                          & 1224.42                          \\
		RD                                                 & BiLSTM                                               & 0.31                            & 0.33                             & 0.79                            & 0.84                             & 0.43                            & 0.43                             & 0.89                            & 0.89                             & 964.62                          & 1662.68                          \\
		RD                                                 & RoBERTa                                             & 0.18                            & 0.21                             & 0.41                            & 0.48                             & 0.46                            & 0.45                             & 0.95                            & 0.95                             & 936.76                          & 1569.42  \\       \hline                
	\end{tabular}
\end{table}

As we can see, BAm improves the BODEGA score in many cases due to a higher success score. However, the semantic score has been slightly reduced because this model additionally selects substitute words that are less related with the meaning of the original. Also we can see that BAm model performs more queries.\footnote{The model performs more queries, but not twice as many as would be expected after increasing the number of substitutes from 36 to 72. This is due to the fact that the algorithm has a stop function if it finds a significant substitute that changes the victim's decision and does not has to check (all) other substitutes in the queue. By contrast, with more substitutes, this extended model has a better chance of changing the victim's mind when the basic version of the model exhausts all the words without a success.}

In order to improve the results, we decided to provide different method of selecting important words for replacement. Comparison from initial version to modified version is shown below:

\begin{itemize}
	\item In \textbf{initial approach} for each word in the sentence (text), the word is replaced with a ''[UNK]'' token (unknown token), creating multiple masked versions of the sentence. So, each masked version has a different word replaced by ''[UNK]''. The victim model is used to compute the probabilities for each of these masked sentences. For each word, an importance score is calculated based on the change in probabilities to an opposite class (label) caused by masking that word. The words are ranked based on their importance scores in descending order. This ranking indicates the significance of each word in contributing to the classification decision of the victim model.
	\item In \textbf{our approach} for each masked word in sentence (text) we try to use each of substitute (36 candidates at the first iteration) and find which of them maximizes the gap (difference between the original and new prediction probabilities). Next we created the list of those initially masked words sorted by the ability to change a decision of the victim model based on the maximum gap score. 
\end{itemize}

\begin{figure}[!ht]
	\centering
	\includegraphics[width=1\linewidth]{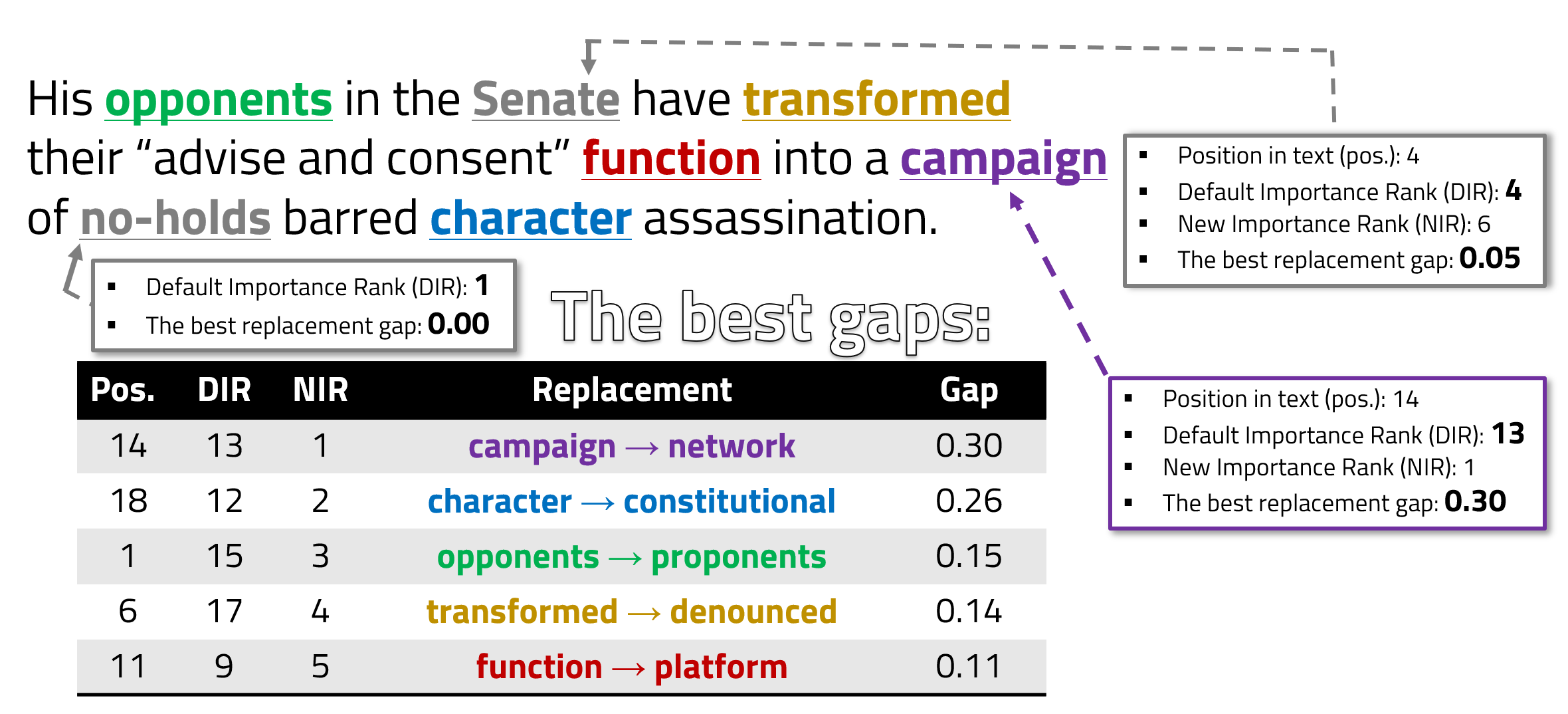}
	\caption{The most important words selected by our approach on the example of the 8th sample from PR2 dataset. Source: own work.}
	\label{fig:pr2-sample8}
\end{figure}

As we could expect, the new importance ranking of the words for replacement affects the results. Now model can choose other word(s) for replacement based on the potential that can be done based on particular candidates for substitution. The figure \ref{fig:pr2-sample8} presents the most important words selected by our approach on the example of the 8th sample from PR2 dataset. We can observe differences between Default Important Rank (DIR) of words in sentence (text) given by initial approach and New Importance Rank (NIR) given by our approach. Please note, that the numbering of word positions ('Pos.' in the figure \ref{fig:pr2-sample8}) in the considered sentence starts from 0.

After selecting the most important word in sentence (text) we replace it with substitute, which gives the biggest positive (for us at this iteration) change in the probabilities of the victim model. This iteration is called ''0th iteration''. It is important to note, here we used also filter for substitutes (which excludes some words from the list of candidates) how it was proposed in initial approach (for example, stop words as it was proposed in \cite{li2020bertattack}). If at this iteration the victim model did not change the decision (the word replacement did not affects classification result) then we go to next iteration (''1st''), which have following modification comparing to 0th:
\begin{itemize}
	\item two most important words in the text (instead of one) are replaced by the best substitutes;
	\item there is no filter for substitutes anymore.
\end{itemize}	

If the 0th and 1st iteration did not succeed, then our approach go through additional iterations from 2nd to 5th. Each of them increased word to be replaced by 1 in the sentence (text) selected from the most important ones. So in the 2nd, 3rd, 4th and 5th iteration it will replace 3, 4, 5, 6 and 7 words in the sentence in total respectively. Additionally, at the last ''5th iteration'' we additionally added common punctuation and digits to the list of potential substitutes to increase the chances of success.

The original paper describing the BERT-Attack showed that using different number of substitutes (candidates) affects the attack success rate \cite{li2020bertattack}. Generally larger number of candidates gives better success rate. In our first experiments we use default number of candidates - 36. We use this number for 0th and 1st iteration. At each subsequent iteration $i$ from 2nd to 5th, we increased the number $n$ of substitutes base on equation: $n(i) = i*36$.

Above described iterations are performed one by one and if a certain iteration succeed the attack it will also stop the process. Those (additional to BAm) modifications of BERT-Attack are named as \textbf{BAm2}. Additionally in cases when after all iterations the victim's decision cannot be changed, we ran another attacker - Genetic \cite{alzantot2018generating} with twice-increased number of population size (40 instead of 20 in the default implementation). In such cases (which failed using BAm2 method), Genetic algorithm was able to perform effective text replacements due to its ability to effectively explore a diverse set of potential solutions through evolutionary strategies such as selection, crossover, and mutation. This approach allows the algorithm to avoid getting trapped in local optima, a common issue in other optimization algorithms. By maintaining a diverse population of candidate solutions and iteratively refining them, the Genetic algorithm can efficiently navigate the search space and identify high-quality solutions that might be missed by more deterministic or less flexible algorithms. 

Those additional (second) modifications of BERT-Attack algorithm with Genetic algorithm for failed cases we called - \textbf{BAm2\&Genetic}. After mentioned before additional modification we found that the results were improved. For example, on the FC task we observed the following improvements comparing to the previous modified version of BERT-Attack ('BAm', see the table \ref{tab:bam}):

\begin{itemize}
	\item \textbf{BERT victim}. BODEGA score: \textbf{0.66} (instead of 0.54), Success score: \textbf{0.97} (instead of 0.78)
	\item \textbf{BiLSTM victim}. BODEGA score: \textbf{0.71} (instead of 0.62), Success score: \textbf{1.0} (instead of 0.89)
	\item \textbf{RoBERTa victim}. BODEGA score: \textbf{0.70} (instead of 0.59), Success score: \textbf{1.0} (instead of 0.84)
\end{itemize}

Next we decided to try another approach which transforms a given text by replacing its words with synonyms in the word embedding space. It takes prebuilt counter-fitted GLOVE words embedding proposed by \cite{mrkvsic2016counter}. After checking the impact of each substitution of the selected word the approach selects those changes that have positive effect on the result, and after choosing the most significant one it greedily updating the set with the highest scoring transformations. Additionally it has constraints which disallow the modification of words which have already been modified and also which disallow the modification of stopwords. Let's name this method as \textbf{GSWSE} (Greedy Search with Word Swap by Embedding).

The GSWSE algorithm also was unable to change the model decision in some cases. So we decided to ran TextFooler method for such texts. This \textbf{GSWSE\&TextFooler} (or GSWSE\&TF) method gave additional improvements of BODEGA score. However, it decreased the success score. For example, on the FC task we observed the following changes in scores comparing BAm2\&Genetic and GSWSE\&TextFooler:

\begin{itemize}
	\item \textbf{BERT victim}. BODEGA score: \textbf{0.74} (instead of 0.66), Success score: 0.94 (instead of \textbf{0.97})
	\item \textbf{BiLSTM victim}. BODEGA score: \textbf{0.75} (instead of 0.71), Success score: 0.93 (instead of \textbf{1.0})
	\item \textbf{RoBERTa victim}. BODEGA score: \textbf{0.71} (instead of 0.70), Success score: 0.93 (instead of \textbf{1.0})
\end{itemize}

In our last stage of experiments we decided to add the CLARE method \cite{li2021contextualized} to our adversarial attackers. This algorithm outperforms other adversarial attack methods due to its unique mask-then-infill procedure using a pre-trained masked language model. This approach allows CLARE to generate fluent and grammatical adversarial examples through three types of contextualized perturbations: Replace, Insert, and Merge. These perturbations provide flexibility in modifying text length and enhance the algorithm's effectiveness in attacking models with fewer edits. CLARE's ability to preserve textual similarity, fluency, and grammaticality while achieving a high attack success rate makes it superior to baseline methods. The \textbf{CLARE} algorithm gave improvements of BODEGA, success and other scores. For example, on the FC task we observed the following changes in scores comparing GSWSE\&TextFooler and CLARE:

\begin{itemize}
	\item \textbf{BERT victim}. BODEGA score: \textbf{0.80} (instead of 0.74), Success score: \textbf{1.0} (instead of 0.94)
	\item \textbf{BiLSTM victim}. BODEGA score: \textbf{0.79} (instead of 0.75), Success score: \textbf{0.98} (instead of 0.93)
	\item \textbf{RoBERTa victim}. BODEGA score: \textbf{0.79} (instead of 0.71), Success score: \textbf{1.0} (instead of 0.93)
\end{itemize}

\section{Results}
\label{results}

The main purpose of the experiment was to test the BODEGA rate with the datasets and to check how the research methods compare with the baseline of a semantic evaluation of the robustness. 
We compared the results with the baseline presented in \cite{przybyła2023verifying}.
The results of HN, FC and RD datasets are from the experiments presented in \cite{przybyła2023verifying}.
In order to compare the results, it was necessary to repeat an experiment on PR2 for establishing baseline.
The analogous problem occurred with C19 -- the dataset needed to be run on our side, as there was no baseline established yet.

Table \ref{tab:baseline} includes results of the attack on BERT and BiLSTM classifiers.

\begin{table}[]
\caption{The baselines and result of adversarial attacks on the BERT and BiLSTM classifier in five misinformation detection tasks. Evaluation measures include
BODEGA score (B.), confusion score (con), semantic score (sem), character score (char)
and number of queries to the attacked model (Q.). The best score in each task and scenario is in boldface.}
\label{tab:baseline}
\centering
\begin{tabular}{|c|c|ccccc|ccccc|}
\hline
\multirow{2}{*}{Task} & \multirow{2}{*}{Method} & \multicolumn{5}{c|}{BERT} & \multicolumn{5}{c|}{BiLSTM} \\
\cline{3-12}
 & & B. & con & sem & char & Q. & B. & con & sem & char & Q. \\
\hline
\multirow{8}{*}{HN} & BAE & 0.34 & 0.60 & 0.58 & 0.96 & 606.83 & 0.48 & 0.77 & 0.64 & 0.98 & 489.27 \\
 & BERT-ATTACK & 0.60 & 0.96 & 0.64 & 0.97 & 648.41 & 0.64 & \textbf{0.98} & 0.66 & 0.99 & 487.85 \\
 & DeepWordBug & 0.22 & 0.29 & 0.78 & 1.00 & 395.94 & 0.41 & 0.53 & 0.77 & \textbf{1.00} & 396.18 \\
 & Genetic & 0.40 & 0.86 & 0.47 & 0.98 & 2713.80 & 0.44 & 0.94 & 0.48 & 0.98 & 2029.31 \\
 & SememePSO & 0.16 & 0.34 & 0.50 & 0.99 & 341.70 & 0.21 & 0.42 & 0.50 & 0.99 & 313.51 \\
 & PWWS & 0.38 & 0.82 & 0.47 & 0.98 & 2070.78 & 0.44 & 0.93 & 0.48 & 0.99 & 2044.96 \\
 & SCPN & 0.00 & 0.92 & 0.08 & 0.02 & 11.84 & 0.00 & 0.94 & 0.08 & 0.02 & 11.89 \\
 & TextFooler & 0.39 & 0.92 & 0.44 & 0.94 & 660.52 & 0.43 & 0.94 & 0.47 & 0.97 & 543.68 \\
 & GSWSE\&TF & \textbf{0.91} & \textbf{1.0} & \textbf{0.92} & \textbf{0.99} & - & \textbf{0.89} & 0.97 & \textbf{0.92} & 0.99 & - \\
 & CLARE & - & - & - & - & - & - & - & - & - & - \\
\hline
\multirow{8}{*}{PR2} & BAE & 0.11 & 0.18 & 0.69 & 0.94 & 312.33 & 0.15 & 0.23 & 0.72 & 0.94 & 32.94 \\
 & BERT-ATTACK & 0.43 & 0.7 & 0.68 & 0.9 & 80.16 & 0.53 & 0.8 & 0.72 & 0.91 & 61.51 \\
 & DeepWordBug & 0.28 & 0.36 & 0.79 & \textbf{0.96} & 27.43 & 0.29 & 0.39 & \textbf{0.79} & \textbf{0.96} & 27.45 \\
 & Genetic & 0.5 & 0.84 & 0.65 & 0.89 & 962.4 & 0.54 & 0.88 & 0.67 & 0.89 & 782.15 \\
 & SememePSO & 0.41 & 0.68 & 0.66 & 0.9 & 96.17 & 0.47 & 0.76 & 0.68 & 0.89 & 85.34 \\
 & PWWS & 0.47 & 0.75 & 0.68 & 0.91 & 131.92 & 0.53 & 0.84 & 0.69 & 0.90 & 130.85 \\
 & SCPN & 0.09 & 0.46 & 0.36 & 0.46 & 11.46 & 0.12 & 0.55 & 0.39 & 0.50 & 11.55 \\
 & TextFooler & 0.43 & 0.77 & 0.64 & 0.87 & 57.94 & 0.51 & 0.85 & 0.67 & 0.88 & 52.59 \\
 & GSWSE\&TF & 0.64 & 0.95 & 0.75 & 0.89 & - & 0.65 & 0.93 & 0.76 & 0.9 & - \\
 & CLARE & \textbf{0.68} & \textbf{0.97} &  \textbf{0.77} & 0.89 & - & \textbf{0.65} & \textbf{0.94} & 0.77 & 0.89 & - \\
\hline
\multirow{8}{*}{FC} & BAE & 0.34 & 0.51 & 0.70 & 0.96 & 80.69 & 0.36 & 0.55 & 0.69 & 0.96 & 77.76 \\
 & BERT-ATTACK & 0.53 & 0.77 & 0.73 & 0.96 & 146.73 & 0.60 & 0.86 & 0.73 & 0.95 & 132.80 \\
 & DeepWordBug & 0.44 & 0.53 & \textbf{0.84} & \textbf{0.98} & 54.32 & 0.48 & 0.58 & \textbf{0.85} & \textbf{0.98} & 54.36 \\
 & Genetic & 0.52 & 0.79 & 0.70 & 0.95 & 1215.19 & 0.61 & 0.90 & 0.71 & 0.95 & 840.99 \\
 & SememePSO & 0.44 & 0.64 & 0.71 & 0.96 & 148.20 & 0.53 & 0.76 & 0.72 & 0.96 & 112.84 \\
 & PWWS & 0.48 & 0.69 & 0.72 & 0.96 & 225.27 & 0.57 & 0.82 & 0.73 & 0.96 & 221.60 \\
 & SCPN & 0.09 & 0.90 & 0.29 & 0.31 & 11.90 & 0.08 & 0.75 & 0.29 & 0.32 & 11.75 \\
 & TextFooler & 0.46 & 0.70 & 0.70 & 0.93 & 106.13 & 0.55 & 0.82 & 0.71 & 0.94 & 98.31 \\
 & GSWSE\&TF & 0.74 & 0.95 & 0.81 & 0.96 & - & 0.75 & 0.93 & 0.84 & 0.97 & - \\
 & CLARE & \textbf{0.8} & \textbf{1.0} & 0.83 & 0.97 & - & \textbf{0.8} & \textbf{0.98} & 0.84 & 0.97 & - \\
\hline
\multirow{8}{*}{RD} & BAE & 0.07 & 0.18 & 0.41 & 0.98 & 313.01 & 0.09 & 0.21 & 0.43 & 0.98 & 312.77 \\
 & BERT-ATTACK & 0.18 & 0.44 & 0.43 & 0.96 & 774.31 & 0.29 & 0.79 & 0.41 & 0.89 & 985.52 \\
 & DeepWordBug & 0.16 & 0.23 & 0.70 & \textbf{0.99} & 232.74 & 0.16 & 0.24 & 0.68 & \textbf{0.99} & 232.75 \\
 & Genetic & 0.20 & 0.46 & 0.45 & 0.96 & 4425.11 & 0.32 & 0.71 & 0.47 & 0.96 & 3150.24 \\
 & SememePSO & 0.10 & 0.21 & 0.46 & 0.97 & 345.89 & 0.15 & 0.31 & 0.48 & 0.97 & 314.63 \\
 & PWWS & 0.16 & 0.38 & 0.45 & 0.97 & 1105.92 & 0.29 & 0.64 & 0.47 & 0.97 & 1059.07 \\
 & SCPN & 0.01 & 0.38 & 0.16 & 0.10 & 11.35 & 0.01 & 0.55 & 0.17 & 0.09 & 11.53 \\
 & TextFooler & 0.16 & 0.41 & 0.43 & 0.91 & 657.15 & 0.24 & 0.64 & 0.41 & 0.87 & 639.97 \\
 & GSWSE\&TF & \textbf{0.64} & \textbf{0.78} & \textbf{0.86} & 0.95 & - & \textbf{0.84} & \textbf{0.95} & \textbf{0.9} & 0.98 & - \\
 & CLARE & - & - & - & - & - & - & - & - & - & - \\
\hline
\multirow{8}{*}{C19} & BAE & 0.16 & 0.3 & 0.56 & 0.94 & 77.21 & 0.24 & 0.42 & 0.59 & 0.95 & 74.75 \\
 & BERT-ATTACK & 0.42 & 0.74 & 0.6 & 0.95 & 161.7 & 0.5 & 0.85 & 0.61 & 0.95 & 127.1 \\
 & DeepWordBug & 0.28 & 0.39 & 0.72 & \textbf{0.99} & 61.06 & 0.34 & 0.48 & 0.72 & \textbf{0.99} & 61.14 \\
 & Genetic & 0.39 & 0.82 & 0.52 & 0.92 & 1562.16 & 0.47 & \textbf{0.92} & 0.54 & 0.94 & 1040.76 \\
 & SememePSO & 0.24 & 0.47 & 0.54 & 0.95 & 228.45 & 0.32 & 0.6 & 0.56 & 0.96 & 184.37 \\
 & PWWS & 0.32 & 0.64 & 0.53 & 0.92 & 276.82 & 0.44 & 0.84 & 0.55 & 0.95 & 268.2 \\
 & SCPN & 0.06 & 0.82 & 0.25 & 0.26 & 11.82 & 0.05 & 0.83 & 0.24 & 0.25 & 11.83 \\
 & TextFooler & 0.35 & 0.72 & 0.53 & 0.92 & 28.49 & 0.42 & 0.85 & 0.53 & 0.92 & 115.78 \\
 & GSWSE\&TF & 0.66 & 0.85 & 0.82 & 0.95 & - & 0.72 & 0.88 & \textbf{0.85} & 0.96 & - \\
 & CLARE & \textbf{0.73} & \textbf{0.91} & \textbf{0.83} & 0.96 & - & \textbf{0.73} & 0.91 & 0.84 & 0.96 & - \\
\hline
\end{tabular}
\end{table}
\FloatBarrier

In table \ref{tab:baseline} the method CLARE is presented only for the tasks PR2, FC and C19.
The remaining tasks were skipped for the method due to the lack of time.
The process of running this method was costly in time and resources.
Additionally, we experienced difficulties with setting the number of queries to the attacked model for CLARE and GSWSE\&TF;
as a result, this parameter was skipped.

The hyperpartisan news detection task (HN) remained the easiest one;
for the victim BERT, GSWSE\&TF achieved 0.91 BODEGA score, which is the best value in the attacks against BERT.
This high value was possible due to the confusion score on level 1.0, which is the maximum level.
The other parameters -- the semantic score and the character score -- remained the highest values for attacking the BERT victim.
The attack against the BiLSTM victim exhibits almost the same properties. 
GSWSE\&TF has the highest scores apart from character score and confusion score, which is on the level of 0.99 and 0.98; only the method DeepWordBug yielded a better result in character score and BERT-ATTACK in confusion score.

The propaganda recognition task (PR2) significantly differs from the HN task in terms of the text length.
The PR2 dataset contains rows with more sentences than HN.
This is because the former consists of full articles, while the latter contains instances of isolated sentences. 
As a result, the PR2 task is less likely to achieve high semantics score than the HN task.
The highest BODEGA score is on the level of 0.68 for BERT and 0.65 for BiLSTM when the attack is carried out using the CLARE method.
These are one of the lowest BODEGA scores in all the tasks excluding the RD task in the BERT victim.
As mentioned earlier, the cause of the lower BODEGA score values is mostly the low semantic score in both victims. 

The BODEGA score of the fact-checking task (FC) reached the level of 0.8 for both victim models: BERT and BiLSTM. 
Both of these values were achieved due to the high values of the confusion score and the character score of that task.
While the confusion score hit the maximum level of 1.0 for the BERT victim and achieved a relativly high level for BiLSTM (0.98),
the character score achieved the level of 0.97.
Both of the researched methods -- GSWSE\&TF and CLARE -- achieved significantly better results than the baseline methods. 

The rumour detection task (RD) was considered by \cite{przybyła2023verifying} as the hardest problem to attack which is further indicated by the lowest BODEGA score when attacking the BERT victim. 
GSWSE\&TF improved the result more than three times compared to the baseline when attacking the BERT victim and improves almost four times when attacking BiLSTM.
The best score in the confusion score category was achieved by the GSWSE\&TF method, which is 0.78.
The best sementics score was achieved by the same method and the highest score in the character score category was achived by DeepWordBug. 
The highest BODEGA score was achived by GSWSE\&TF method.
For the BiLMST victim, the BODEGA score is 0.84,
achieved by GSWSE\&TF and at the same time the highest scores for the confusion and the semantic categories are achieved by the same method. 
In the character score, DeepWordBug achieved the highest value -- 0.99; however, as demonstrated earlier, GSWSE\&TF is only marginally lower than that, i.e. 0.98.

Finally, the covid-19 task (C19) was solved on the level of 0.73 BODEGA score for both victims -- BERT and BiLSTM. 
Achieving high BODEGA score attacking the BERT victim was done by CLARE, which was the best method in the confusion score (0.91) and the semantic score (0.83)
and was only slightly worse than the best method in the character score category -- DeepWordBug -- which accomplished 0.99, whereas CLARE achived 0.96. 
On the other hand, the highest scores were achieved by different methods while attacking BiLMST:
the confusion score -- Genetic with the value on the level of 0.92; the semantic score -- GSWSE\&TF (0.85); the character score -- DeepWordBug (0.99).
However, it is worth to note that in all these cases CLARE was marginally worse, so it amounted to best BODEGA score. 

\begin{table}[]
\caption{The results of adversarial attacks on the RoBERTa classifier in five misinformation detection tasks with methods GSWSE\&TF and CLARE. Evaluation measures include
BODEGA score (B.), confusion score (con), semantic score (sem), character score (char)
and number of queries to the attacked model (Q.). The best score in each task and scenario is in boldface.}
\label{tab:finalresults}
\centering
\begin{tabular}{|c|c|ccccc|}
\hline
\multirow{2}{*}{Task} & \multirow{2}{*}{Method} & \multicolumn{5}{c|}{RoBERTa} \\
\cline{3-7}
 & & B. & con & sem & char & Q. \\
\hline
\multirow{2}{*}{HN} & GSWSE\&TF & 0.83 & 0.99 & 0.86 & 0.97 & - \\
 & CLARE & - & - & - & - & - \\
\hline
\multirow{2}{*}{PR2} & GSWSE\&TF & 0.49 & 0.82 & 0.68 & 0.85 & - \\
 & CLARE & 0.62 & 0.93 & 0.75 & 0.87 & - \\
\hline
\multirow{2}{*}{FC} & GSWSE\&TF & 0.71 & 0.93 & 0.8 & 0.95 & - \\
 & CLARE & 0.8 & 1.0 & 0.82 & 0.97 & - \\
\hline
\multirow{2}{*}{RD} & GSWSE\&TF & 0.55 & 0.71 & 0.83 & 0.93 & - \\
 & CLARE & - & - & - & - & - \\
\hline
\multirow{2}{*}{C19} & GSWSE\&TF & 0.67 & 0.97 & 0.75 & 0.92 & - \\
 & CLARE & 0.73 & 0.99 & 0.79 & 0.93 & - \\
\hline
\end{tabular}
\end{table}

Table \ref{tab:finalresults} inculdes results of the attack on RoBERTa classifier. 
RoBERTa classifier was made available later than BERT and BiLSTM, which was mentioned ealier.
The baseline wasn't established and we found difficulties to establish them by ourself because of lack of time. 
Testing the RoBERTa classifier were the most time and resources consuming testing from all classifiers.
RoBERTa was a surprise classifier and it was used to evaluate our methods. 

In table \ref{tab:finalresults} are presented results of two methods -- GSWSE\&TF and CLARE.
Both of this methods ere the fundation of are research.
Because lack of the time and resources we skipped the CLARE methods on the HN and RD tasks.
As mention before we struggled with setting the number of queries to the attack model, so in the table 
this parameters are skkiped. 

We observed higher values on BODEGA score using the CLARE method. 
But at the same time the BODEGA scores are lower than the BODEGA scores achieved testing our methods with the BERT and BiLSTM victims in tasks HN, PR2 and RD.
In rest of the task -- FC and C19 -- we achieved the same BODEGA scores like with BERT and BiLSTM. 

We can summerise the experiments through the following general observations: 

\begin{itemize}
\item We found high semantic score the most difficult to achieve.
\item The larger the text, the more difficult it is to achieve high BODEGA score.
\item We found RoBERTa classifier as the most difficult to achieve high the BODEGA score.
\end{itemize}

\subsection{Qualitative Analysis of Manual Evaluation Results}

The CheckThat! Lab Task 6 is founded on the assumption that adversarial examples can alter classification results without changing the message's underlying meaning. To support automated verification, a manual annotation procedure was implemented. Adversarial examples were manually classified into three categories:
\begin{itemize}
\item (a) Preserve the Semantic Meaning,
\item (b) Change the Semantic Meaning,
\item (c) No Sense.
\end{itemize}

A significant discrepancy was observed between automatically generated semantic scores and manual verification results. Automatic scores ranged from 0.68 to 0.86, with CLARE consistently yielding higher results. In contrast, manual scoring indicated that only 0.11 of adversarial examples preserved the semantic meaning. We believe this discrepancy arises from two main sources.

Firstly, while BLEURT aligns well with human judgment, its sensitivity may be lower compared to human annotators, despite its ability to handle multi-word modifications and contradictions.

Secondly, BLEURT processes long text fragments by splitting them into sentences and averaging the semantic similarities between sentence pairs. This approach differs significantly from manual scoring, which evaluates adversarial examples as a whole, resulting in much lower scores.

Upon receiving guidelines for manual annotation, we conducted a qualitative analysis of a subset of adversarial examples generated by our methods. We found that approximately half of these examples introduced incorrect information not present in the original text (e.g., 'Teck was a professional athlete.' changed to 'Teck was a professional maid.'). About a quarter of the changes were negations of the original text (e.g., 'Grease did not have a soundtrack.' changed to 'Grease did include a soundtrack.'). Another 10\% of examples failed to preserve the semantic meaning for various reasons, such as loss or addition of information, occasional French words substituting English, or pure nonsense. Only about 15\% of successful perturbations fully preserved the semantic meaning (e.g. 'Stay safe when you return to work.' changed to 'staying safe when you return to work.'). This analysis pertains to single perturbations. The datasets contained 2,231 examples, comprising over 13,000 sentences with an average of 6 sentences per example. This implies that most of the semantic meaning of the complete text was preserved, even when the semantic meaning of a single sentence in a multi-sentence text was altered.

\section{Conclusions and Future Works}
\label{conclusions}
In this study, we explored the application of ensemble learning to the domain of adversarial attacks on text classifiers. Our approach was motivated by the principle that leveraging multiple adversarial attack methods can yield better performance compared to any single method. Through systematic experimentation and modification of existing attack algorithms, we aimed to improve the effectiveness and efficiency of adversarial attacks against various victim models.

We began with a literature review and replicated some of the prominent adversarial attack methods, using implementations from the OpenAttack library. Initial experiments indicated that BERT-Attack and Genetic Algorithm were generally effective, though other methods like TextFooler, DeepWordBug, and SCPN also showed promise in specific scenarios. This led us to test these methods on our five datasets: C19, FC, HN, PR2, and RD, initially targeting two victim models: BERT and BiLSTM.

Our first set of modifications to BERT-Attack (resulting in the BAm variant) involved adjusting parameters to enhance the balance between semantic preservation and attack success rate. These adjustments improved performance across several metrics, as shown in our comparative analysis.

Further, we introduced a novel method for ranking word importance, which led to the BAm2 variant. This method evaluates the impact of potential word substitutions on model predictions, thereby allowing more informed and effective attacks. Additionally, for cases where BAm2 failed, we integrated the Genetic Algorithm to bolster the attack success rate. This combination, BAm2\&Genetic, yielded significant improvements, particularly in the BODEGA and success scores.

We then explored an alternative approach, GSWSE, which employs word embeddings for synonym replacement. This method, coupled with TextFooler for the challenging cases, offered further enhancements in BODEGA score, albeit with a slight reduction in the success rate.

In the final stage of our experiments, we used the CLARE method. CLARE's innovative mask-then-infill procedure, leveraging a pre-trained masked language model, demonstrated superior performance across all metrics. It provided a high attack success rate while maintaining textual similarity and fluency, making it the most effective method in our ensemble.

Finally, the developed solutions were tested on the third victim model: RoBERTa. The evaluation confirmed the effectiveness of our approach in  generating adversarial text in five problem domains.

Our findings underscore the potential of ensemble learning in the adversarial attack domain. By combining multiple attack strategies and continuously refining our methods, we achieved notable improvements in attack effectiveness. Future work could explore additional ensemble configurations and further optimizations, aiming to develop even more robust adversarial attack frameworks.

In future works we plan to extend the methods by using various data sources and large language models (LLMs). For example, as a comprehensive text corpus, Wikipedia can be used to train or fine-tune language models for better contextual understanding and generation. Using Wikipedia articles, the quality of adversarial examples can be further enhanced by providing more natural and contextually appropriate substitutions, particularly when we can consider quality differences between Wikipedia articles in various language versions \cite{lewoniewski2023companies}. Especially, we plan to use publicly available services with measures related to information quality and reliability of sources, such as BestRef \cite{bestref}, WikiRank \cite{wikirank} and various revscoring models \cite{LiftWing}. Future research can leverage the analysis of Wikipedia references to enhance adversarial attacks by incorporating high-quality scientific sources into training datasets, ensuring more credible and contextually relevant adversarial examples \cite{lewoniewski2023understanding}.

Such structured knowledge bases as Wikidata and DBpedia can provide rich semantic context and relationships between entities. Integrating those and other data sources \cite{hellmann2021towards} could help in refine word importance ranking and improve the selection of substitutes by ensuring semantic consistency and relevance. For example, entity linking and relationship extraction from these knowledge bases can ensure that replacements preserve factual accuracy and context.

The advancements in adversarial attack strategies can be further developed by integrating insights from the analysis of large language models (LLMs) like ChatGPT on the fake news phenomenon \cite{wecel2023}. Utilizing state-of-the-art and novel LLMs, such as Claude, Gemini, GPT-4, LLama, Mistral, can improve the quality of generated adversarial examples. Fine-tuning these models on specific datasets or incorporating them into the ensemble can lead to more sophisticated and effective attacks. LLMs can also be used to simulate human-like understanding and generation of text, making adversarial examples more natural and harder to detect. Moreover, combining traditional adversarial attack methods with the capabilities of LLMs can create hybrid approaches that leverage the strengths of both. For example, initial perturbations can be generated using classical methods, followed by refinement and enhancement using LLMs.

\begin{acknowledgments}
	The research is supported by the project ``OpenFact -- artificial intelligence tools for verification of veracity of information sources and fake news detection'' (INFOSTRATEG-I/0035/2021-00), granted within the INFOSTRATEG I program of the National Center for Research and Development, under the topic: Verifying information sources and detecting fake news.
\end{acknowledgments}

\bibliography{biblio}

\end{document}